%% file: main.tex
\documentclass[letterpaper, 10 pt, conference]{ieeeconf}
\IEEEoverridecommandlockouts    
\usepackage{glossaries}

\input{glossary}

\let\labelindent\relax

\usepackage[table,usenames,dvipsnames]{xcolor}      
\usepackage{extarrows}                              
\usepackage{enumitem}
\usepackage{wrapfig}

\usepackage{multirow, multicol}
\usepackage{array}
\newcolumntype{P}[1]{>{\centering\arraybackslash}p{#1}}
\newcolumntype{M}[1]{>{\centering\arraybackslash}m{#1}}
\usepackage{booktabs}
\newcolumntype{N}{>{\centering\arraybackslash}m{.5in}}
\newcolumntype{G}{>{\centering\arraybackslash}m{2in}}

\usepackage{amsmath,amssymb,amsfonts,amsthm,dsfont} 
\usepackage{algorithm, algorithmicx, listings}        
\usepackage[noend]{algpseudocode}			              
\makeatletter
\def\BState{\State\hskip-\ALG@thistlm}
\makeatother

\usepackage{graphicx}
\usepackage{tabularx}
\usepackage{subcaption}
\usepackage{textcomp}

\usepackage[font={small}]{caption}   
\usepackage[breaklinks=true, colorlinks, bookmarks=true, citecolor=Black, urlcolor=Violet,linkcolor=Black]{hyperref}

\usepackage{svg}

\usepackage{amssymb,graphicx}

\newcolumntype{C}[1]{>{\centering\arraybackslash}p{#1}}

\usepackage{everypage}
\usepackage{afterpage}
\usepackage{ifthen}

\usepackage[normalem]{ulem}
\usepackage[utf8]{inputenc}
\usepackage[english]{babel}
\usepackage{hyperref}
\hypersetup{
    colorlinks=true,
    linkcolor=blue,
    filecolor=magenta,      
    urlcolor=cyan,
}

\DeclareMathAlphabet\mathbfcal{OMS}{cmsy}{b}{n}

\newtheorem*{assumption*}{Assumption}

\newtheorem*{problem*}{Problem}

\usepackage{lipsum}
\usepackage[capitalise]{cleveref}
\usepackage{svg}
\usepackage{mathtools}

\begin{document}
\title{\LARGE \bf Spatio-Temporal Metric-Semantic Mapping for Persistent Orchard Monitoring: Method and Dataset}

\author{
Jiuzhou Lei, Ankit Prabhu, Xu Liu, Fernando Cladera, Mehrad Mortazavi, \\ Reza Ehsani, Pratik Chaudhari, Vijay Kumar
\thanks{
We gratefully acknowledge the support of the IoT4Ag Engineering Research Center funded by the National Science Foundation (NSF) under NSF Cooperative Agreement Number EEC-1941529, ARL DCIST CRA W911NF-17-2-0181, NSF Grant CCR-2112665, NIFA grant 2022-67021-36856, and NVIDIA. 
 We thank Prof. Drew Wilkerson, and the owners of \textit{Hands on Earth} and \textit{Shaw} Orchards for their support in data collection.
}
\thanks{J. Lei, A. Prabhu, F. Cladera, P. Chaudhari, and V. Kumar are with the GRASP Laboratory, University of Pennsylvania, Philadelphia, PA, 19104, USA {\tt\small\{jiuzl, praankit, fclad, pratikac, kumar\}@seas.upenn.edu}. X. Liu is with the Department of Aeronautics and Astronautics, Stanford University, Stanford, CA 94305, USA (e-mail: {\tt\small xuliu@stanford.edu}). M. Mortazavi and R. Ehsani are with the Department of Mechanical Engineering, University of California, Merced, CA, 95343, USA {\tt\small\{smortazavi3, rehsani\}@ucmerced.edu}.} 
\thanks{This work has been submitted to the IEEE for possible publication. Copyright may be transferred without notice, after which this version may no longer be accessible.}
}

\maketitle

\begin{abstract}
Monitoring orchards at the individual tree or fruit level throughout the growth season is crucial for plant phenotyping and horticultural resource optimization, such as chemical use and yield estimation. We present a 4D spatio-temporal metric-semantic mapping system that integrates multi-session measurements to track fruit growth over time. Our approach combines a LiDAR-RGB fusion module for 3D fruit localization with a 4D fruit association method leveraging positional, visual, and topology information for improved data association precision. Evaluated on real orchard data, our method achieves a 96.9\% fruit counting accuracy for 1,790 apples across 60 trees, a mean fruit size estimation error of 1.1 cm, and a 23.7\% improvement in 4D data association precision over baselines. We publicly release a multimodal dataset covering five fruit species across their growth seasons. \url{https://4d-metric-semantic-mapping.org/}
\end{abstract}



\input{tex/introduction}

\input{tex/related-work}

\input{tex/proposed-approach.tex}
\input{tex/results-and-analysis.tex}

\input{tex/conclusion.tex}

\bibliographystyle{IEEEtran}
\bibliography{ref}

\end{document}

%% file: glossary.tex
\newacronym{mav}{UAV}{Unmanned Aerial Vehicle}
\newacronym{uav}{UAV}{Unmanned Aerial Vehicle}
\newacronym{ovc}{OVC}{Open Vision Computer}
\newacronym{lidar}{LiDAR}{Light Detection and Ranging}
\newacronym{vio}{VIO}{visual-inertial odometry}
\newacronym{satnav}{GNSS}{Global Navigation Satellite Systems}
\newacronym{dgps}{DGPS}{Differential GPS}
\newacronym{rtk}{RTK}{Real-Time Kinematics}
\newacronym{ppk}{PPK}{Post-Processed Kinematic}
\newacronym{gpgpu}{GPGPU}{General-Purpose Graphics Processing Unit}
\newacronym{hri}{HRI}{Human-Robot Interaction}
\newacronym{ugv}{UGV}{Unmanned Ground Vehicle}
\newacronym{uwb}{UWB}{Ultra Wideband}
\newacronym{svm}{SVM}{Support Vector Machine}
\newacronym{fcn}{FCN}{Fully Convolutional Network}
\newacronym{cnn}{CNN}{Convolutional Neural Network}
\newacronym{loam}{LOAM}{LiDAR Odometry and Mapping}
\newacronym{sloam}{SLOAM}{Semantic LiDAR Odometry and Mapping}
\newacronym{slam}{SLAM}{Simultaneous Localization and Mapping}
\newacronym{swap}{SWaP}{Size, Weight, and Power}
\newacronym{iot4ag}{IoT4Ag}{NSF Engineering Research Center for the Internet of Things for Precision Agriculture}
\newacronym{grasp-lab}{GRASP Lab}{the General Robotics, Automation, Sensing and Perception Laboratory}
\newacronym{jps}{JPS}{Jump Point Search}
\newacronym{ukf}{UKF}{Unscented Kalman Filter}
\newacronym{sam}{SAM}{Smoothing and Mapping}
\newacronym{icp}{ICP}{Iterative Closest Point}
\newacronym{imu}{IMU}{Inertial Measurement Unit}
\newacronym{tsdf}{TSDF}{Truncated Signed Distance Field}
\newacronym{esdf}{ESDF}{Euclidean Signed Distance Field}
\newacronym{rrt}{RRT}{A rapidly exploring random tree}
\newacronym{mslam}{MS-SLAM}{Metric-Semantic SLAM}
\newacronym{iou}{IoU}{Intersection over Union}

%% file: tex/introduction.tex
\section{Introduction}
\label{sec:introduction}

\begin{figure}[ht!]
    \centering
    \includegraphics[width=0.8\columnwidth]{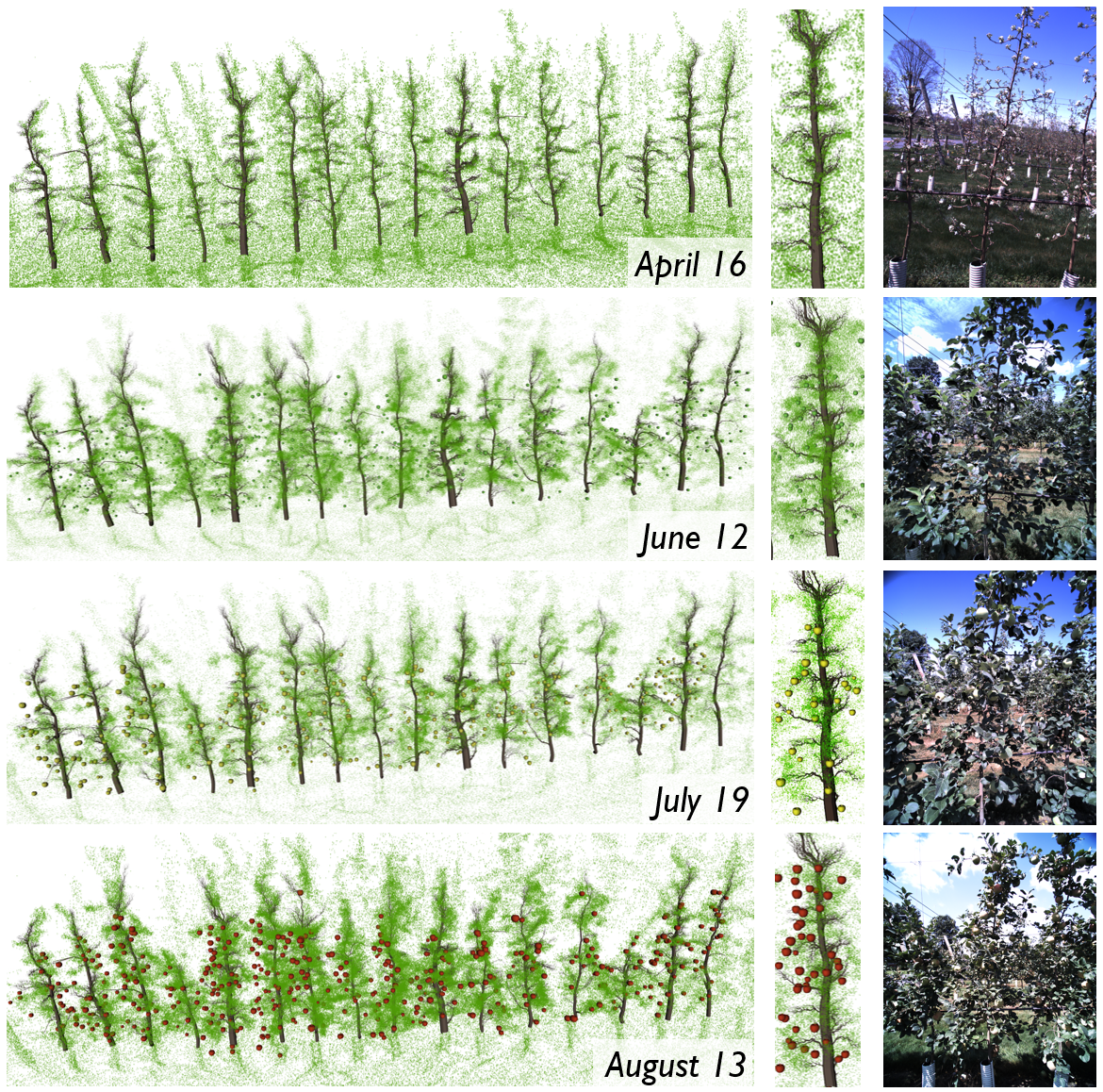}
    \caption{\textit{A 4D metric-semantic map of an apple orchard from April to August. Each row represents a time session. The first panel shows the metric-semantic map.
    The middle panel zooms in on the 6th tree, while the right panel presents its raw RGB image. Tree skeleton meshes \cite{adtree} are for visualization only.}}
\label{fig:title-fig-4d-mapping}
\end{figure}

Persistent orchard monitoring is vital for farmers to optimize cultivation and maximize yield. By monitoring fruit growth, farmers can make informed decisions, such as applying chemical thinning based on fruit size, adjusting irrigation according to growth rates, and optimizing labor for harvest through yield estimation.
Spatio-temporal (4D) mapping of an orchard can also help enhance plant phenotyping \cite{fiorani2013future, watt2020phenotyping} by monitoring important traits of plants and providing insights into how they change over the entire growth season. 
Hence, this motivates the problem of 4D orchard mapping, which in this context is defined as 3D orchard mapping with temporal information. The actionable information obtained from 4D orchard mapping is that each fruit is associated with a timestamp, and its state is monitored across its growth season. 
In our 4D metric-semantic map, the semantics involve detecting the presence of fruits, while metrics such as fruits' counts, sizes, and positions can be obtained.

While previous works, like \cite{chen2017counting, bargoti2017image, liu2018robust}, have demonstrated impressive accuracy in fruit localization for a single time session, they fall short in monitoring fruit growth over time. Extending 3D fruit tracking to a 4D monitoring framework requires data association of fruits across time, which remains a challenge due to their positional changes and variations in appearance. Existing methods, such as point cloud registration \cite{carlone2015towards}, feature-based image registration \cite{dong20174d, robust-long-term-registration}, and hybrid 2D-3D approaches \cite{4d_plant_growth_2023}, have shown success in tracking plant growth but lack the granularity needed for individual fruit instances.

To address the challenges in 4D metric-semantic mapping in orchards, we developed a 3D fruit localization module taking LiDAR points and RGB images for fruit localization and absolute size estimation. 
A two-stage data association algorithm leveraging positional, visual, and topological information is proposed to associate fruits from different time sessions under temporal changes. 
In addition, a dataset is collected from real-world, natural orchards and made available for public access during our experiments. The dataset, containing high-resolution RGB images and LiDAR point clouds, could facilitate other branches of agricultural research in small fruit detection\cite{smallFruitDetection}, flower counting\cite{flowerDetection}, tree branch modeling\cite{AppletreeModeling}.
Unlike most existing datasets such as \cite{fuji-air, fuji-sfm, bargoti2017deep, MinneApple, ag-survey}, which typically capture only a single growth stage, our dataset spans the entire growth cycle, from the bud stage through to harvest. 

\begin{figure*}[ht!]
        \centering          
            \includegraphics[trim=0 0 0 0, clip, width=0.95\textwidth]{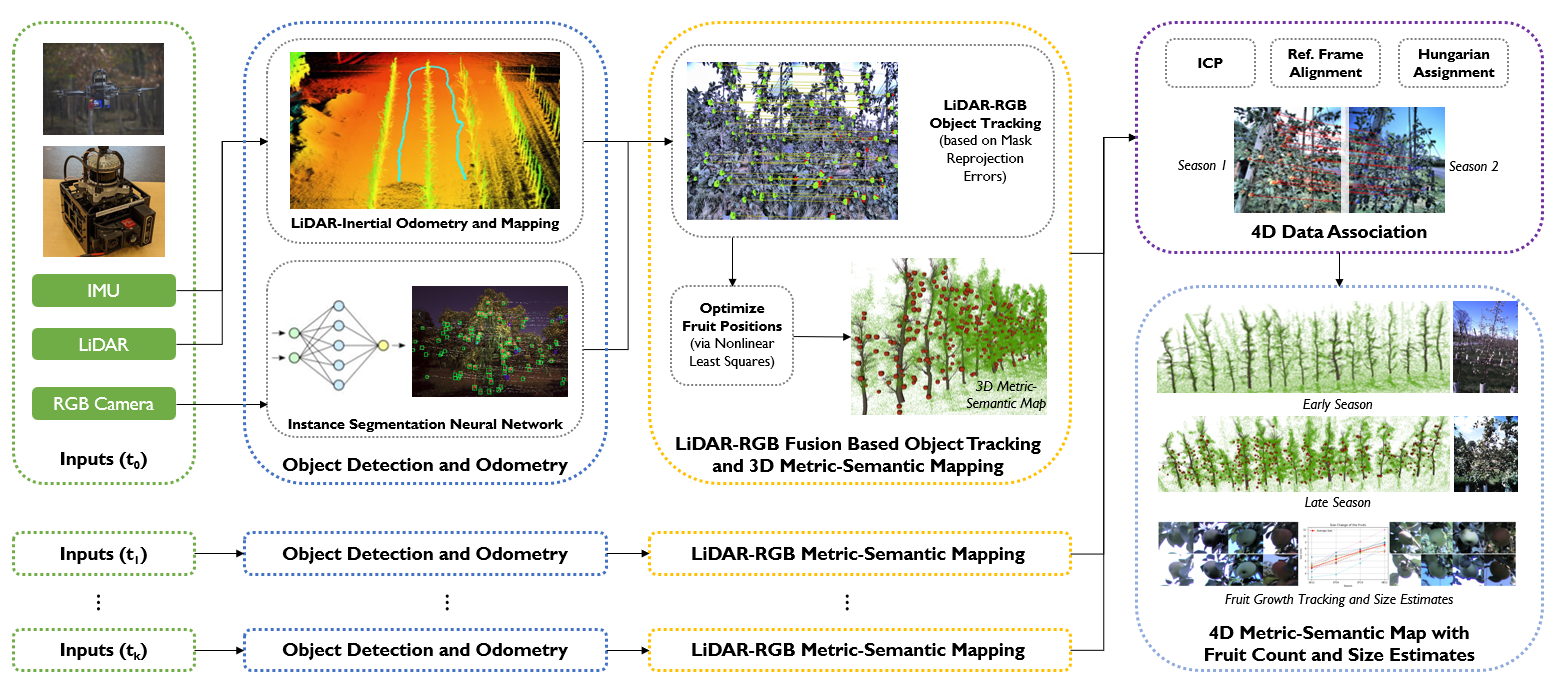}
    \vspace{-0.1in}
        \caption{\textit{System Diagram.} \textit{\uline{Module 1}} (green box): Our system takes in sensor data from the LiDAR, RGB camera, and \gls{imu}. \textit{\uline{Module 2}} (blue box): \textit{Object detection and odometry}. First, we use {Faster-LIO} \cite{bai2022faster}, a LiDAR-inertial odometry (LIO) algorithm for pose estimation and point cloud motion undistortion. Meanwhile, {YOLO-v8}, an instance segmentation model \cite{Jocher_YOLOv8_by_Ultralytics_2023}, is used on the RGB images to detect and segment fruits. \textit{\uline{Module 3}} (orange box): \textit{LiDAR-RGB fusion for fruit localization}. The point cloud is back-projected into the segmentation mask to estimate the depth of a fruit. Fruit detections from different image frames are tracked using the Hungarian assignment algorithm with mask IoU as the cost.
        Then, we minimize the reprojection error of fruit centroids to optimize the fruit positions. 
        \textit{\uline{Module 4}} (purple box): 4D Data Association  takes optimized 3D fruit landmarks and images as input. 
        Fruits are associated across sessions using our two-stage matching algorithm, with a cost function based on position, visual, and topology information.
        \textit{\uline{Module 5}} (cyan box): \textit{4D metric-semantic map generation}. Using the 4D data association, we can construct a 4D metric-semantic map, acquiring actionable information such as fruit counts, sizes, and positions throughout the entire growth season.
        }
        \label{fig:4d-mapping-system-diagram}
    \vspace{-0.2in}
\end{figure*}

Our \textbf{contributions} can be summarized as follows:
\begin{enumerate}
    \item We develop a mapping framework that takes in measurements across multiple time sessions and constructs a 4D metric-semantic map, along with the estimation of the count and size of fruits, bridging the gap between 3D fruit localization and 4D fruit growth monitoring. 
    \item We introduce a 4D fruit association algorithm for tracking fruits throughout their growth season, achieving performance that significantly surpasses baseline methods.
    \item We publicly release a multimodal dataset capturing LiDAR point clouds, RGB images, GPS, and IMU measurements across an entire growth season for five different fruits, complementing existing datasets with extended temporal coverage.
\end{enumerate}

%% file: tex/related-work.tex
\section{Related Work}
\textbf{3D Fruit Localization and Counting.}
To enable 4D spatio-temporal data association of fruits, an accurate measurement of fruits from a single time session is crucial. In \cite{chen2017counting, bargoti2017image}, deep learning algorithms were used to directly count fruits from images and obtain a yield estimate. 
In \cite{liu2018robust}, fruits were tracked frame by frame robustly using a combination of the Hungarian Assignment algorithm, Kalman Filter, optical flow \cite{Lucas:1981:IIR:1623264.1623280}, and Structure for Motion (SfM) algorithm for data association and outlier rejection. 
To count fruits from both sides of a row of trees and accelerate computation, Liu et al. \cite{liu2019monocular} used the fruit instances as features to replace SIFT \cite{lowe2004distinctive} features, which directly constructed a map of fruits and estimated the camera trajectory.
Meyer et al. \cite{meyer2024fruitnerfunifiedneuralradiance} modeled an orchard using NeRF, taking into consideration fruit instance information, from which the fruits can be counted directly in 3D space and localized. 
Although these works focus on fruit localization and counting during a single time session and demonstrate impressive fruit-counting accuracy, they fall short of monitoring fruits across multiple time sessions. Our work aims to bridge the gap between 3D fruit localization and 4D fruit monitoring.
 
\textbf{Spatial-Temporal Mapping for Agriculture.}
The key component extending 3D mapping to 4D mapping is the temporal (4D) data association. In \cite{carlone2015towards}, Carlone et al. registered the point clouds of crops via expectation-maximization. Dong et al. in \cite{dong20174d} and Nived et al. in \cite{robust-long-term-registration} both proposed to extract 2D image features that are robust to seasonal variation for image registration. \cite{4d_plant_growth_2023} combined both 2D and 3D visual features to find temporal correspondences. In more recent works, \cite{segmentation_4d_plants_2020} and \cite{non-rigid-registration-3d-points-cloud} proposed to match the skeletons or key points extracted from high-precision point clouds of plants for growth tracking and phenotyping. While these approaches successfully monitor changes in a plot of crops or individual plants, they lack the granularity needed in tracking fruit instances that are subject to occlusions from canopy and changes in position, size, and appearance. 

As for applications in fruits, Riccardi et al. in \cite{fruit_tracking_hign_precisoion_cloud} used terrestrial laser scanners to obtain colored and precise fruit point clouds. For each fruit instance, a descriptor encoding both positions and neighboring fruit layouts was designed for fruit matching. 
Fusaro et al. in \cite{fusaro2024horticultural} took colored and high-resolution point clouds as input for fruit instance segmentation. For each fruit instance, the point clouds within a range of the fruit instance were used to extract descriptors with Minkowski convolutional neural networks \cite{minkowskiNet}. The extracted descriptors were then used as input to an attention-based matching network for temporal fruit association.
However, fruit segmentation on point clouds is subject to extensive labeling. Additionally, acquiring high-precision point clouds can be slow compared to sensors such as RGB cameras and lower-density LiDARs. 
Freeman and Kantor \cite{freeman2025transformerbasedspatiotemporalassociationapple} proposed a transformer-based model for associating apple fruitlets over time using stereo images. Given two stereo images of the same fruitlet cluster at different times, shape and positional descriptors are extracted from each fruitlet’s point cloud using a encoder-decoder architecture based on MinkowskiEngine \cite{minkowskiNet}. A transformer encoder then refines these features, enabling fruitlet association based on feature scores and matchability. However, this approach assumes prior knowledge of fruitlet-cluster assignments and focuses solely on temporal association. 

When integrating 3D fruit localization with 4D fruit association, challenges arise, including duplicated fruit counts, missed detections, occlusion of previously seen fruits, and the emergence of new fruits between time sessions. Our work aims to achieve temporal fruit association under those outlier conditions. Unlike learning-based methods for 4D fruit association, our approach leverages pre-trained models and heuristically derived features, eliminating the need for neural network training.

%% file: tex/proposed-approach.tex

\section{Proposed Approach}
\label{sec:proposed-approach}
The objective is to monitor the growth of fruits over time, given the point cloud and images collected during the ego-motion of a robot or sensor platform. The objective is divided into solving two sub-problems: fruit localization in 3D space and temporal tracking across multiple time sessions, which range from weeks to months. We use time sessions to refer to different times from June to August when the data were collected.

In this section, we describe each module of our framework for 4D metric-semantic mapping. A visualization of the framework is shown in \cref{fig:4d-mapping-system-diagram}.

\subsection{Spatial (3D) Localization} 
\label{subsec:proposed-approach-3D}

3D localization of fruits aims to associate fruit detections between image frames and estimate their 3D positions in each time session. We take the following steps:

\subsubsection{Fruit Instance Point Cloud Extraction}
\label{subsubsec:3D-tracking-inital-depth}
To obtain point clouds belonging to each fruit instance, we use LiDAR for depth estimation and RGB images for fruit instance segmentation. A fine-tuned YOLOv8 \cite{Jocher_YOLOv8_by_Ultralytics_2023} is used for instance segmentation to obtain the instance mask of each fruit. 
Faster-LIO \cite{faster-lio} is used to estimate the pose of the sensor. 
Our LiDAR and RGB camera are time synchronized, and their extrinsic (rigid body transformation) and intrinsic parameters are calibrated before the experiments. 
Based on such information, each image's corresponding camera pose is obtained. 
We then project the point clouds onto the image plane to obtain each fruit instance's point cloud. 
However, the point cloud from a single scan is sparse. To remedy this, we accumulate point clouds for 1.5 seconds in our implementation. The position of each fruit is then simply the centroid of the point cloud projected inside the segmentation mask. 

\begin{figure*}[ht!]
        \centering          
            \includegraphics[trim=0 0 0 0, clip, width=0.725\textwidth]{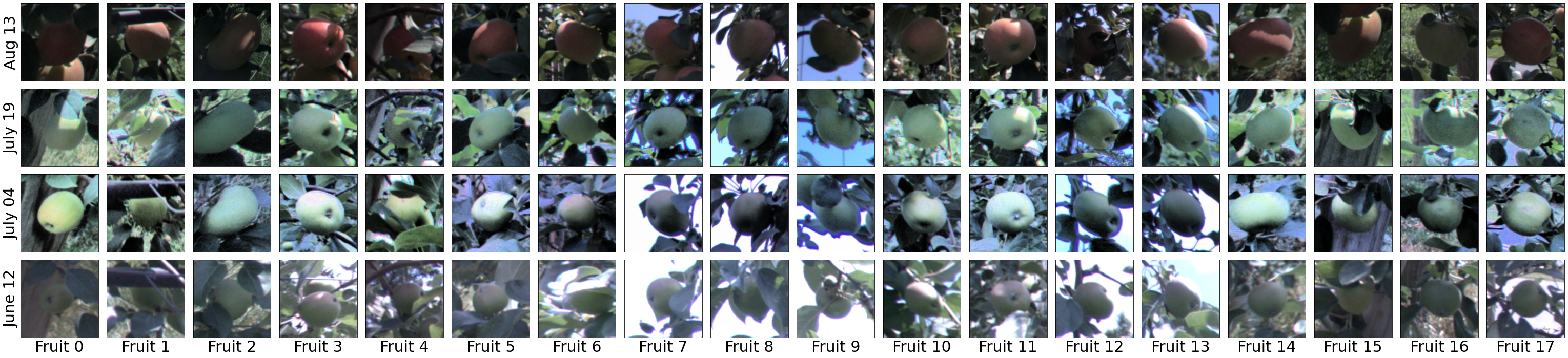}
    ~~
        \centering       
            \includegraphics[trim=0 0 0 20, clip, width=0.25\textwidth]{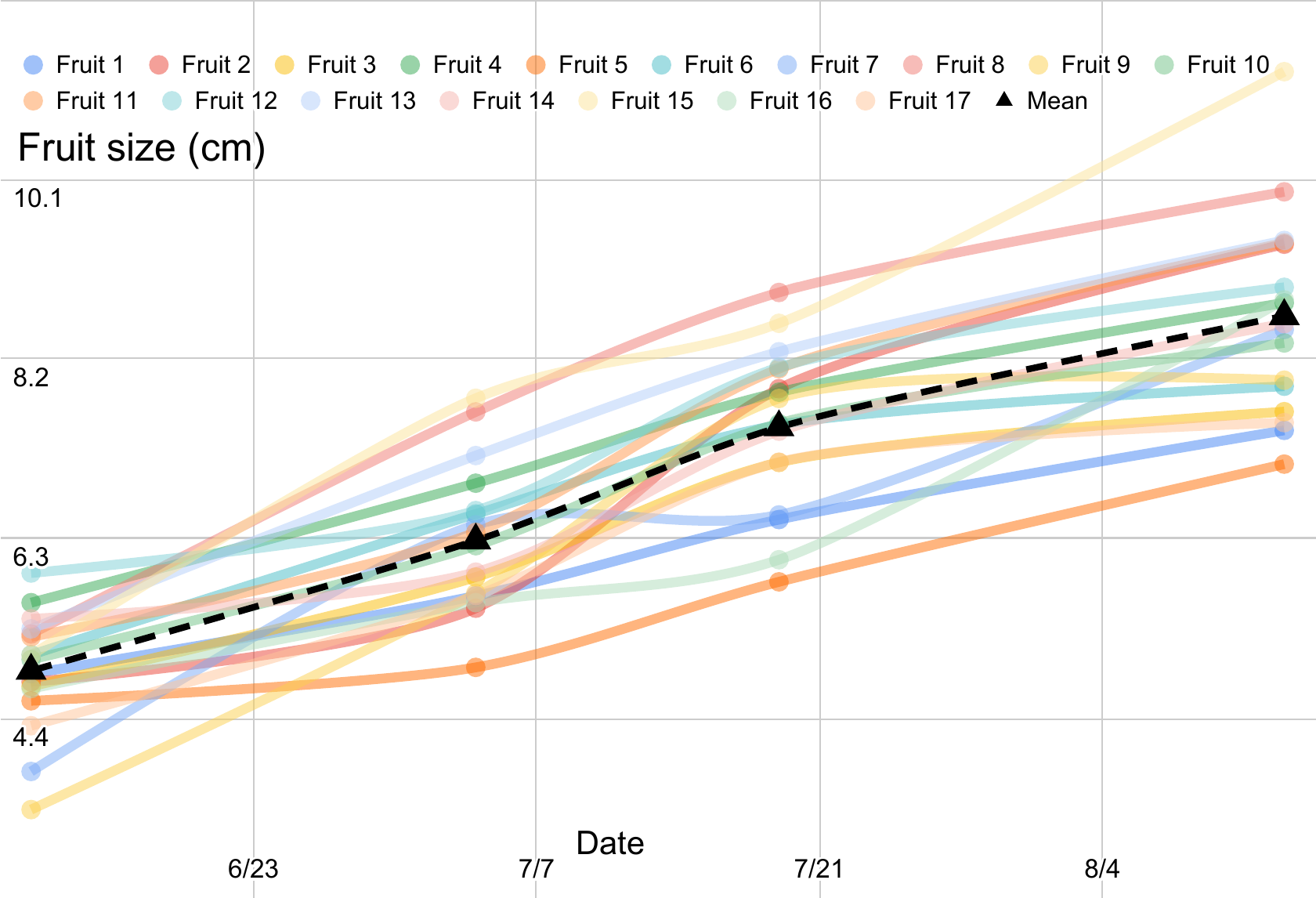}
        \caption{\textit{Precise monitoring of fruit growth}. \uline{\textit{The left panel}} shows examples of our 4D spatio-temporal fruit tracking results. \uline{\textit{The right panel}} quantitatively shows how these fruits grow over time. The \textit{X-axis} shows the date, and the \textit{Y-axis} shows fruit sizes in $cm$. Different colors represent different fruits. Fruit sizes increase over time, with varying but generally similar growth rates reflected by the slope of their trendlines. The average fruit sizes are shown in a black dashed line with triangle markers. 
        Our method automatically generates fruit tracks across multiple seasons, providing users with detailed information on each fruit, such as size, color, shape, and position over time. 
        }
    \vspace{-0.2in}
        \label{subfig:fruit-size-tracking-subfigure fruit size tracking}
        \label{fig:growth_vis}
\end{figure*}

\subsubsection{Tracking across Image Frames}
\label{subsubsec:3D-tracking-between-image-frames}

Let fruits from two consecutive image frames be $\bold{I}_l = \{o_i, i =1,2,...,m\}$ and $ \bold{I}_{l+1} = \{o_j, j = 1, 2, ..., n\}$. To match fruits across consecutive image frames, we seek to find an injective mapping between the two sets of fruits that minimizes the matching cost $C(\bold{I}_l, \bold{I}_{l+1})$. This matching problem can be solved using the Hungarian assignment algorithm. Given the availability of each fruit's depth in the camera frame and the corresponding camera poses, we compute the matching cost using the mask \gls{iou} as follows:

\begin{equation}
\begin{aligned}
    mask_{i, l+1} = proj(T_{l}^{l+1}, p_{i,l}) 
\end{aligned}
\label{eq: mask cost}
\end{equation}

\begin{equation}
\begin{aligned}
    c(o_{i,l}, o_{j,l+1}) = 1 - IoU(mask_{i, l+1}, mask_{j, l+1})
\end{aligned}
\label{eq: mask reprojection}
\end{equation}
$p_{i,l}$ refers to the point cloud of fruit $i$ detected at frame $l$, $T_{l}^{l+1}$ is the relative transformation from frame $l$ to frame $l+1$, $mask_{i,l+1}$ is the image coordinates of $mask_i$ at frame $l+1$, and $proj()$ projects 3D points into image frames.
$c()$ calculates the assignment cost between each association pair. 
To take into account the fact that not all fruits masks at frame $l$ can be assigned to a fruit mask at the next frame (example in the event of a false-negative detection), the cost matrix consists of two parts: $C_{IoU} \in \mathbb{R}^{m \times n}$ with $C_{IoU,ij} = c(o_{i,l}, o_{j,l+1})$ and $C_u \in \mathbb{R}^{m \times m}$ in which all element equals to $1 - u$. $u$ is the minimal \gls{iou} threshold to consider two masks matched as one. $C_u$ allows a fruit not to be matched with any fruit detected in the next frame.

\begin{equation}
\begin{aligned}
    C(I_l, I_{l+1}) = [C_{IoU} \quad  C_u] 
\end{aligned}
\label{eq: cost matrix}
\end{equation}

From this step, we obtain a set of tracks. For clarity, a ``track" refers to a sequence of associated fruit detections across multiple image frames. To avoid overcounting due to potential false positive detections, we only count the fruits that have been observed in more than three consecutive frames. 

\subsubsection{Re-projection Error Minimization}
\label{subsubsec:3D-tracking-projection-error-minimization}
There are multiple sources of errors in the aforementioned steps. The fruit instance point clouds may include noisy points that do not belong to the fruit, and different parts of the fruit are observed from multiple views, leading to small variations in position estimation. 
Hence, we minimize the reprojection error of fruit positions across all image frames to improve depth estimation. Given $n$ camera poses and $m$ fruit positions, we solve the following nonlinear least squares problem to obtain optimized fruit positions and camera poses.

\begin{equation}
\begin{aligned}
\min_{X, R, t} \quad & \sum_{i=1}^{n} \sum_{j=1}^{m} a_{ij}(x_{ij} - proj(R_i, t_i, X_j))^2\\
\textrm{s.t.} \quad & X_{j,0}-d \leq X_j \leq X_{j,0}+d \\
\end{aligned}
\end{equation}
where $x_{ij}$ is the $j$th fruit's pixel coordinates in the $i$th image frame, $X_j$ is the $j$th fruit position, $R_i$ and $t_i$ represents the rotational and translational components of the $i$th camera pose respectively. The binary variable $a_{ij}$ indicates whether fruit $j$ is present in frame $i$. We also constrain the problem such that $X_j$ is within a certain distance $d$ from its initial value $X_{j, 0}$ to avoid optimized positions being too far away from the initial position due to outliers. 

Because of canopy occlusions and missing detections from the instance segmentation model, some fruits may be initially tracked, lost, and then re-tracked again in later frames, resulting in double counts of the same fruit. It is thus necessary to re-identify such previously tracked fruits. Therefore, for each track and its nearby tracks within a certain distance threshold, we project their 3D point cloud into the same image plane. Two conditions must then be met to re-associate the two tracks as one: (1) \gls{iou} between the two masks is higher than a predefined threshold. (2) The detections in each track must occur in distinct image frames. More formally, for any two tracks, $ \bold{Track}_i $ and $ \bold{Track}_j $, representing sequences of image frame indices where the same fruit is detected, $ \bold{Track}_i \cap \bold{Track}_j = \emptyset$, for $ i \neq j$.

The 3D localization module will output the fruit counts and their positions with each fruit assigned a unique index.

\subsection{Spatio-temporal (4D) Tracking}
\label{subsec:4D-tracking}
4D tracking aims to find fruit correspondences between two consecutive time sessions. 
Given the output of the 3D localization and the images from two time sessions, we match fruits from different times in two stages. In the first stage, for each image $I_a$ from time session $A$, we find an image $I_b$ from time session $B$ with the most similar view. 
Then Hungarian assignment algorithm with a cost function $Cost = C_{visual} + C_{position}$ considering both visual similarities and Euclidean distance is applied to associate fruits detected in $I_{a}$ and $I_{b}$. 
Each fruit instance $f_i$ from time session $A$ could be associated with one or multiple fruit instances appearing in different image frames from time session $B$. We denote the association results using $M(f_i) = \{ f_{j,b} \}$, where $f_i$ is the fruit index in time session $A$, and $f_{j,b}$ is the fruit $f_j$'s index in time session $B$ and it appears in image $I_b$. (eg. $M(f_i = 2) = \{ f_{j=0,b=0}=0, f_{j=0, b=1}=0,  f_{j=1, b=1}=1\}$). Fruits with the same subscript $j$ mean they are the same fruit instance, but we add subscript $b$ to indicate that they are associated with $f_i$ when $f_j$ is detected in image $I_b$.
A fruit instance in time session $A$ may be associated with different fruits in different image frames in time session $B$ due to missing detections (caused by canopy occlusion or the instance segmentation model failing to detect fruits), duplicated fruit counts (resulting from unsuccessful re-tracking in 3D tracking), and temporal position changes.

The purpose of the first stage of matching is to extract a subset of fruits from both time sessions to serve as reference points in the images, which will be used to quantify topological similarities for the next stage of matching.  
These reference points must be reliably and consistently matched across different images, analogous to how SIFT image features are used for feature registration between two images.  
We select reliable reference fruits by thresholding the entropy $H(M(f_i))$ for each fruit instance in time session $A$. Lower entropy means one fruit tends to be associated with another fruit consistently across different views. In the implementation, we set the threshold to be 0.8.

In the second stage, the matching process is repeated except for the cost function which also takes into account the topology information, $Cost = C_{visual} + C_{position} + C_{topology}$. The cost between matched reference fruits will be 0. For each fruit instance $f_i$ in time session $A$, a majority vote is taken on $M(f_i)$ to determine its final matched fruit in time session $B$. Refer to Algorithm \ref{alg: two-stage-match} for the matching process described above. 

\begin{algorithm}
\caption{Two-Stage 4D Data Association}
\begin{algorithmic}[1]
    \Require Images $\bold{I}_A, \bold{I}_B$ and counted fruits $\{ f_i \}, \{ f_j \}$ from two time sessions A and B.
    \State Initialize $M(f_i)$ as an empty dictionary.
    \For{each image $I_a$ in session A}
        \State $I_b \gets \arg\min_{b \in B} ||Dino(I_a) - Dino(\bold{I}_B)||_2 $, to find the image $I_b$ of the most similar view in time session B. 
        \State Apply Hungarian assignment between the fruits $\{f_{i,a}\}$ and $\{f_{j,b}\}$ detected in $I_a$ and $I_b$ using $Cost = C_{visual} + C_{position}$. Add $f_{i}$ as the key and $f_{j,b}$ as the value into $M$.
    \EndFor
    \State Select reliable matches $\{(f_i, most\_common(M(f_i)))\}$ where entropy $H(M(f_i)) < \theta_{thres}$.
    \State Repeat step 1 to 4 using $Cost =  C_{visual} + C_{position} + C_{topology}$
    \State \Return Set of matches $\{(f_i, most\_common(M(f_i)))\}$.
\end{algorithmic}
\label{alg: two-stage-match}
\end{algorithm}

Different sources of information are leveraged in the above-mentioned cost function, and this is done as follows:

\textbf{Position cost:} The position cost is the Euclidean distance between two fruits. However, the fruits positions from different time sessions may not be expressed in the same coordinate frames. We obtain the precise relative transformation between the coordinate frames by aligning the point clouds from two successive temporal sessions using \gls{icp} algorithm \cite{planar_icp}. Fruit positions are transformed into the same reference frame before computing the position cost. 
    
\textbf{Visual cost:} For each fruit detected in image $i_a$, we crop a sub-image surrounding the detected fruit with it in the center. The backbone of DINOv2 \cite{dinov2} is used to extract visual features from the sub-image. The visual feature is a $h\times w \times n$ matrix. The visual cost between fruits $f_i$ and $f_j$ would be 
    \begin{equation}
        \begin{aligned}
            C_{visual}(f_i, f_j) = \| Dino(f_i) - Dino(f_j) \|_2
        \end{aligned}
    \end{equation}
    
\textbf{Topology cost:} Topology information, which refers to the spatial relationship between fruits in the same image frame, also remains consistent under temporal changes. 
As presented in \cite{fruit_tracking_hign_precisoion_cloud, sadeghian2017trackinguntrackablelearningtrack}, this information is encoded by discretizing the space around a target and counting the number of surrounding objects within each discretized cell to construct a histogram descriptor. However, the effectiveness of these features depends on the resolution of the discretization. A resolution that is too coarse may reduce descriptiveness, while an overly fine resolution can introduce noise due to occlusions and temporal fruit position changes.
Instead of discretization, we use detected fruits as the feature points in an image. The topology descriptor comprises $n$ vectors pointing from the target fruit to reference fruits in the image coordinate frame. For two fruits with their corresponding topology descriptors $\Vec{a}_k, \Vec{b}_k, k = 0, 1, 2, ..., n - 1$, 
    
\begin{equation}
    \begin{split}
         C_{topology}(f_i, f_j) = \Big| \Big\{ (\vec{a_k}, \vec{b_k}), \\ 
            \cos^{-1} \left( \frac{\vec{a_k} \cdot \vec{b_k}}
            {\|\vec{a_k}\| \|\vec{b_k}\|} \right) > \theta_{thres} \Big\} \Big|
    \end{split}
\end{equation}
The cost is the number of pairs of vectors between which the angle is larger than a threshold. In the implementation, the angle threshold is set to be $10^\circ$.

For the topology cost and visual cost to be effective, the two images used in the cost computation must be captured from a similar viewpoint. To ensure this, we compare their DinoV2 latent features to identify viewpoint similarity. Additionally, we normalize the position cost and visual cost to bring them to the same scale, while keeping the topology cost unchanged. This allows the topology cost to function as a penalty term, rejecting false matches that violate topology consistency.

\section{4D Multi-Modal Datasets in Orchards}
\label{sec:orchard-datasets}

This section gives more details on the collected data. The data was collected by scanning the sides of a row of fruit trees as shown in the object detection and odometry module of \cref{fig:4d-mapping-system-diagram}. 
It includes LiDAR point clouds, RGB images, RGB-D data, IMU, and GPS measurements collected from April to August 2024 across 284 trees: 120 apple trees, 63 pear trees, 50 peach trees, 34 cherry trees, and 17 pistachio trees.

\textbf{Sensor Suite} Our sensor suite is shown in the leftmost (inputs) panel of \cref{fig:4d-mapping-system-diagram}. It consists of an Ouster OS0-128 LiDAR along with a FLIR Chameleon global-shutter RGB camera 
and an Intel RealSense D435i RGB-Depth (RGB-D) camera. It also consists of a high-quality VectorNav VN-100 IMU and UBlox ZED-F9P for GPS measurements. 
We compute the extrinsic calibration between the LiDAR and the global shutter RGB camera by using an initial estimate from the sensor tower's CAD model. 
We further refine the calibration by manually aligning the projected point cloud with an RGB image using a calibration board.
Such sensors can be deployed on a quadrotor as shown in \cref{fig:4d-mapping-system-diagram}

\begin{figure*}[t!]
        \centering          
            \includegraphics[trim=0 0 0 12, clip, width=0.75\textwidth]{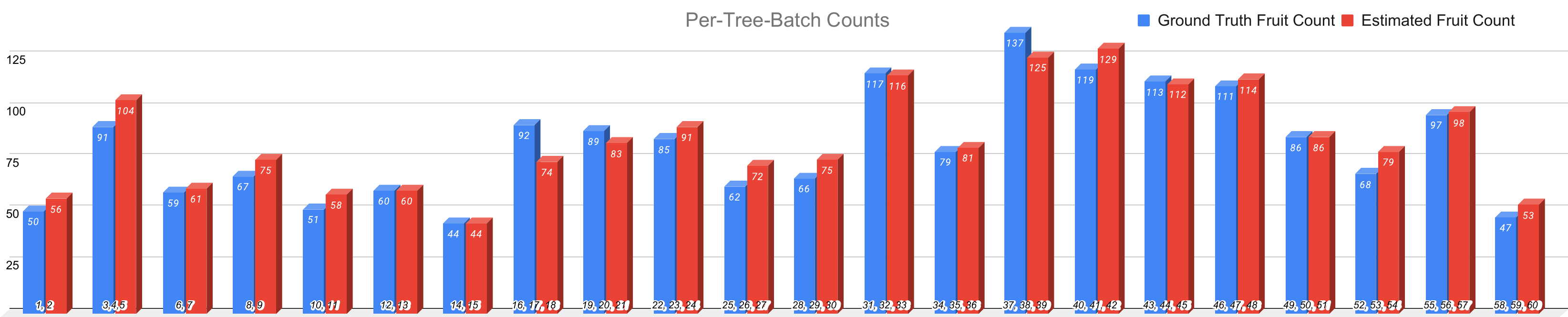}
            
        \caption{\textit{Fruit count ground truth vs. estimated (Y-axis) per tree batch (X-axis).} The \textit{X-axis} represents the IDs of trees included in each batch, with batches containing 2 to 3 trees. For each batch, the estimated counts closely align with the human-labeled ground truth, demonstrating the accuracy of our system. Detailed statistics are provided in \cref{table:apple_count_accuracy}. 
         }
        \label{fig: Per Tree Count}
\end{figure*}

\textbf{Orchard Overview} Data was collected every two or three weeks from the start of the growing season (early April) until late August. 

The types of orchards from which we collected datasets include apple, peach, pear, cherry, and pistachio orchards. The sites where we collected apple and pear datasets are located at the \textit{Hands on Earth Orchard}, Lititz, Pennsylvania, USA. The apple, peach, and cherry datasets were collected from the \textit{Shaw Orchards}, Harford, Maryland, USA. Finally, the pistachio dataset was collected at Merced, California, USA.

%% file: tex/results-and-analysis.tex
\section{Results and Analysis}
\label{sec:results and analysis}

This section presents our results and analysis. 
We evaluate our approach in terms of 3D fruit counting error, 4D fruit association precision, and the accuracy of fruit size estimation for downstream tasks, such as estimating fruit growth rate. 
In each subsection, we first describe how the ground truth data is obtained, followed by an explanation of the evaluation metrics, and conclude with an analysis of the quantitative results.


\subsection{3D Fruit Counting}
\label{subsec:results-3D-tracking}

Accurately obtaining ground truth by counting fruits in an orchard is challenging due to the possibility of double-counting. To address this, fruits are manually counted from images extracted from video streams, aided by digital markers to track counted fruits. However, double-counting may still occur when the same tree appears in multiple frames. While image stitching can create panoramas for counting, it may introduce artifacts that blur fruits. Instead, images, where each tree is fully visible and centered, are selected with careful marking and boundary tracking to prevent duplication.

The 3D localization is evaluated using the counting error. Our 3D fruit counting results are detailed in \cref{table:apple_count_accuracy}. For a total of 60 apple trees, the 3D tracking method gives a 3.1\% total count error. We estimate 1846 apples from our algorithm, and when compared to the ground truth of 1790 apples, it gives us an absolute total count error of 56 apples. Compared to ground truth, our method tends to overestimate counts, primarily due to (1) double counting some fruits due to noisy fruit instance point clouds and (2) counting fruits fallen on the ground, which are excluded when we generate the ground truth. In particular, noisy fruit instance point clouds are obtained because some small and partially occluded fruits cannot accurately be fused from the RGB detections and the point clouds. Hence, they contain noisy background point clouds, which induce errors in their estimated 3D positions.

\cref{fig: Per Tree Count} shows a per tree-batch count estimation compared to the corresponding ground truth. A total of 22 batches, each with two or three trees, was created from 60 apple trees, and their per-batch fruit count was compared with the per-batch ground truth. Trees were separated into these batches based on how easily human annotators can distinguish tree boundaries to generate accurate ground truth counts. We decided to use tree batches for method evaluation instead of individual trees because, when trees are very close together, it is challenging to accurately assign fruits to one or the other.
As seen from \cref{fig: Per Tree Count}, our method consistently estimates fruit counts for trees despite the varying fruit densities. Similarly, the low standard deviation of absolute count error (5 apples) and the small mean of absolute count error (6 apples) shown in \cref{table:apple_count_accuracy} demonstrates the counting accuracy of our method.

\begin{table}[t!]
\begin{center} 
 \setlength\extrarowheight{2.5pt}
 \resizebox{0.482\columnwidth}{!}
 {\begin{tabular}{||c | c ||} 
 \hline
 Stats. (Total Count) & Results \\ [0.6ex] 
 \hline\hline
 Estimated / \textit{Ground Truth} & 1846 / \textit{1790} \\
 \hline\hline
 Absolute (Abs.) Error (Err.)  & 56 / 3.1$\%$ \\
    \hline
\end{tabular}}
~
 \resizebox{0.482\columnwidth}{!}
 {\begin{tabular}{||c | c ||} 
 \hline
 Stats. (Per Tree Batch Counts)  & Results \\ [0.6ex] 
  \hline\hline
 Mean of Abs. Err. & 6.0 / 8.0$\%$ \\
  \hline\hline
Std Dev of Abs. Err.  & 5.0 / 6.4$\%$ \\

    \hline
\end{tabular}}
\end{center}
\caption{\textit{Total count and per-tree-batch count statistics}.
Our method can accurately estimate the yield, as indicated by the 3.1$\%$ error on the total count over 1790 fruits.  
The mean and standard deviation of per-tree-batch counts are calculated from all 22 batches. 
}
   
\label{table:apple_count_accuracy}
\end{table}

\begin{figure}[t!]
        \centering          
            \includegraphics[trim=0 25 0 40, clip, width=0.8\columnwidth]{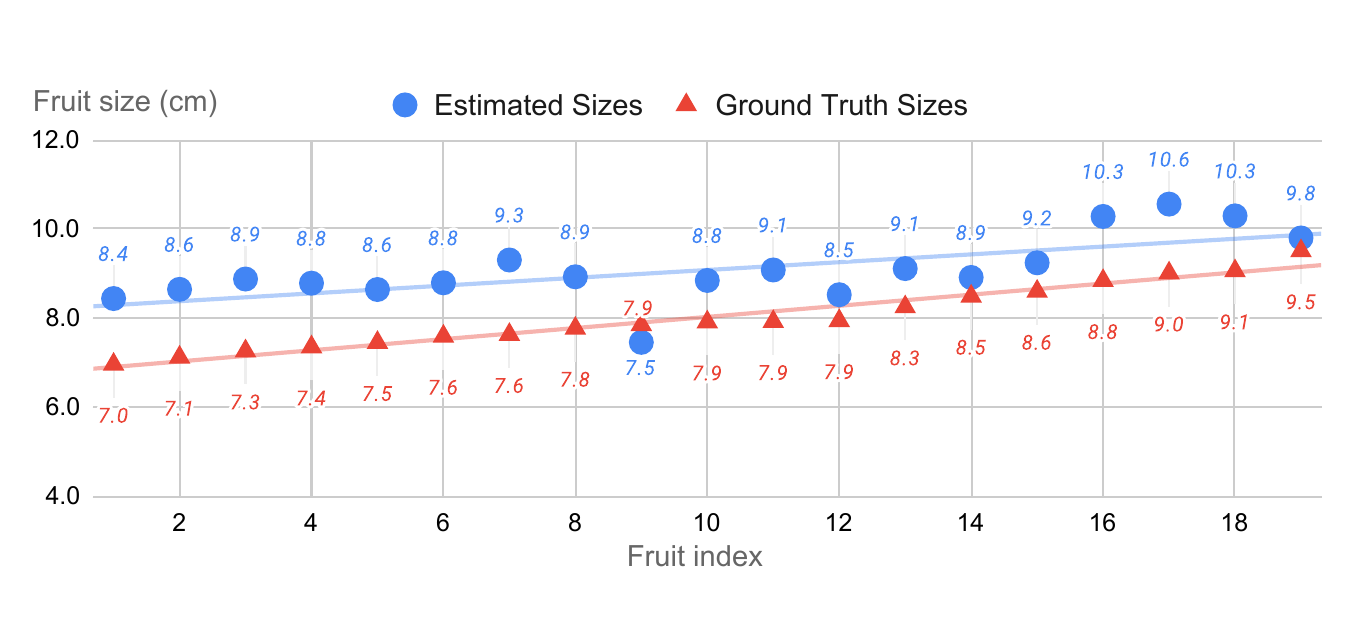}
        \caption{\textit{Fruit size ground truth vs. estimated}. The mean and standard deviation of absolute errors in fruit size estimates are 1.10 $cm$ (0.43 $inch$) and 0.45 $cm$ (0.18 $inch$).}
        \label{fig:size_estimation}
    \vspace{-0.2in}
\end{figure}

\subsection{4D Fruit Association}
\label{subsec:results-4D-tracking}

\begin{figure}[t!]
        \centering          
            \includegraphics[width=0.99\columnwidth]{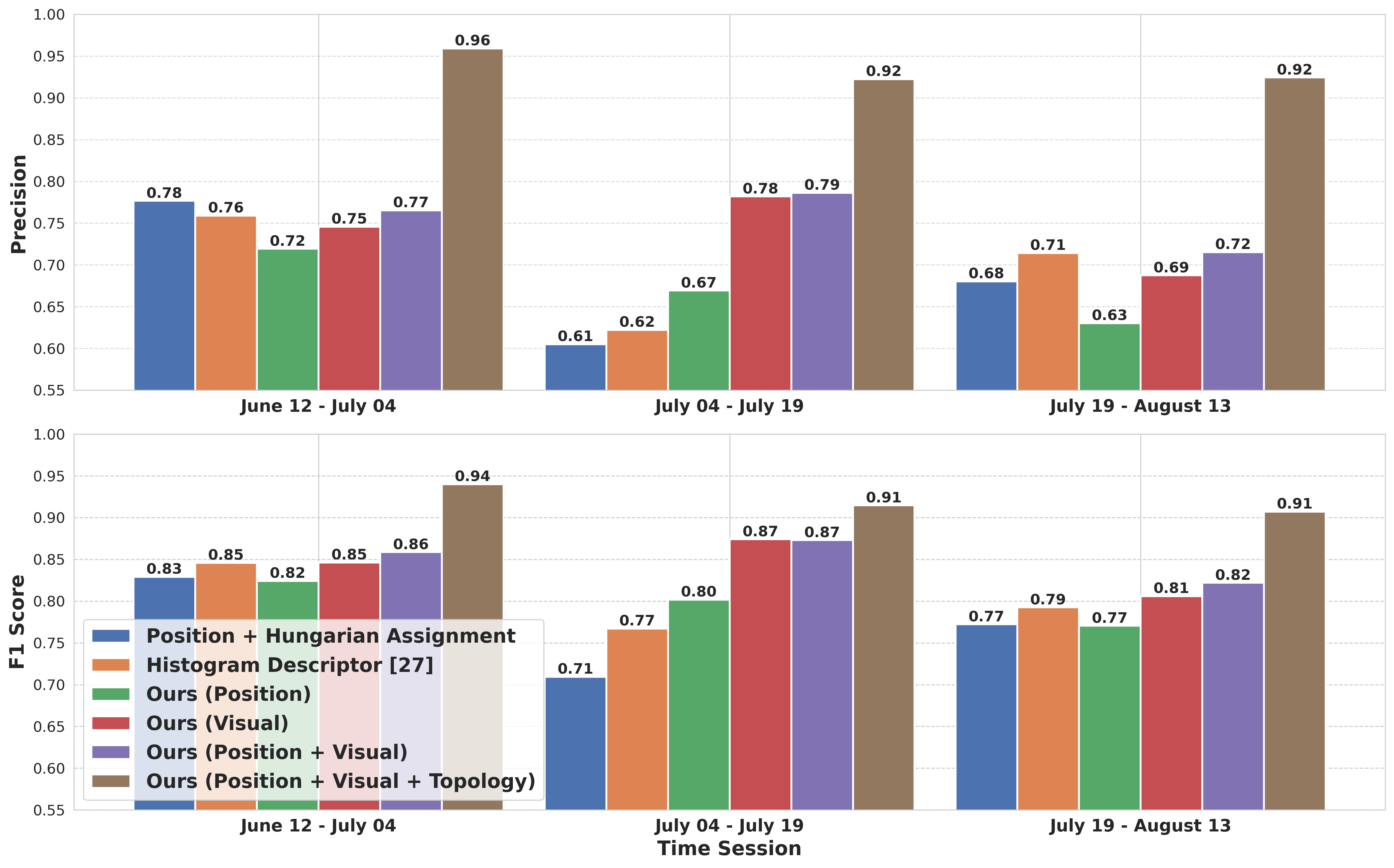}
        \caption{\textit{4D Fruit Association Evaluation in terms of Precision and F1 Score.} From left to right, the first two bars are the results from baselines; the remaining bars represent ablation study results.}
        \label{fig:precision}
\end{figure}

\begin{figure}[t!]
        \centering          
            \includegraphics[width=0.8\columnwidth]{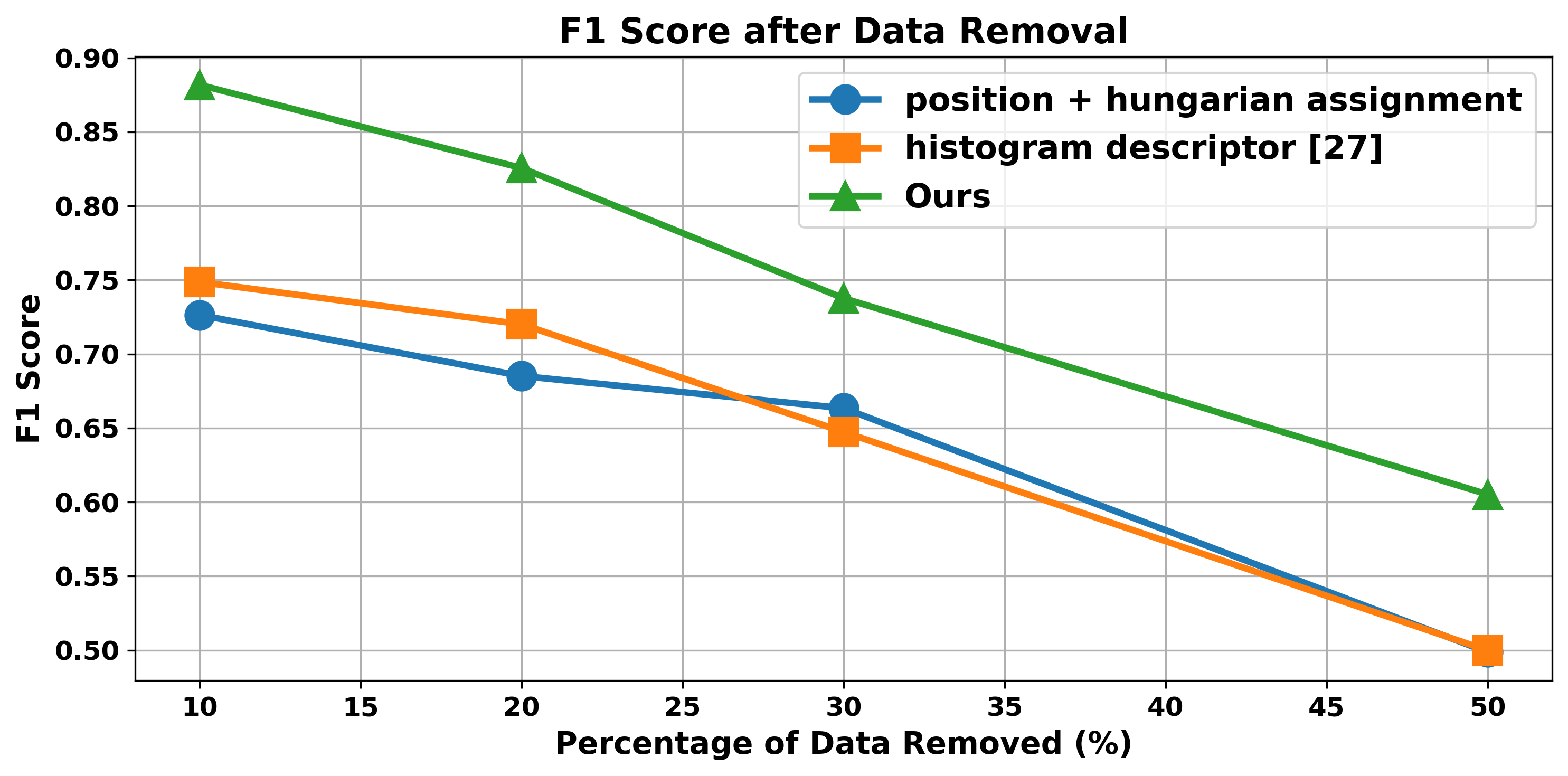}
        \caption{\textit{Effect of removing data on the F1 scores for 4D fruit association.} }
        \label{fig:precision_with_removal}
        \vspace{-0.2in}
\end{figure}

\begin{figure*}[t!]
    \centering
    \begin{subfigure}[t]{0.31\textwidth}
        \centering          
            \includegraphics[ width=1.0\textwidth]{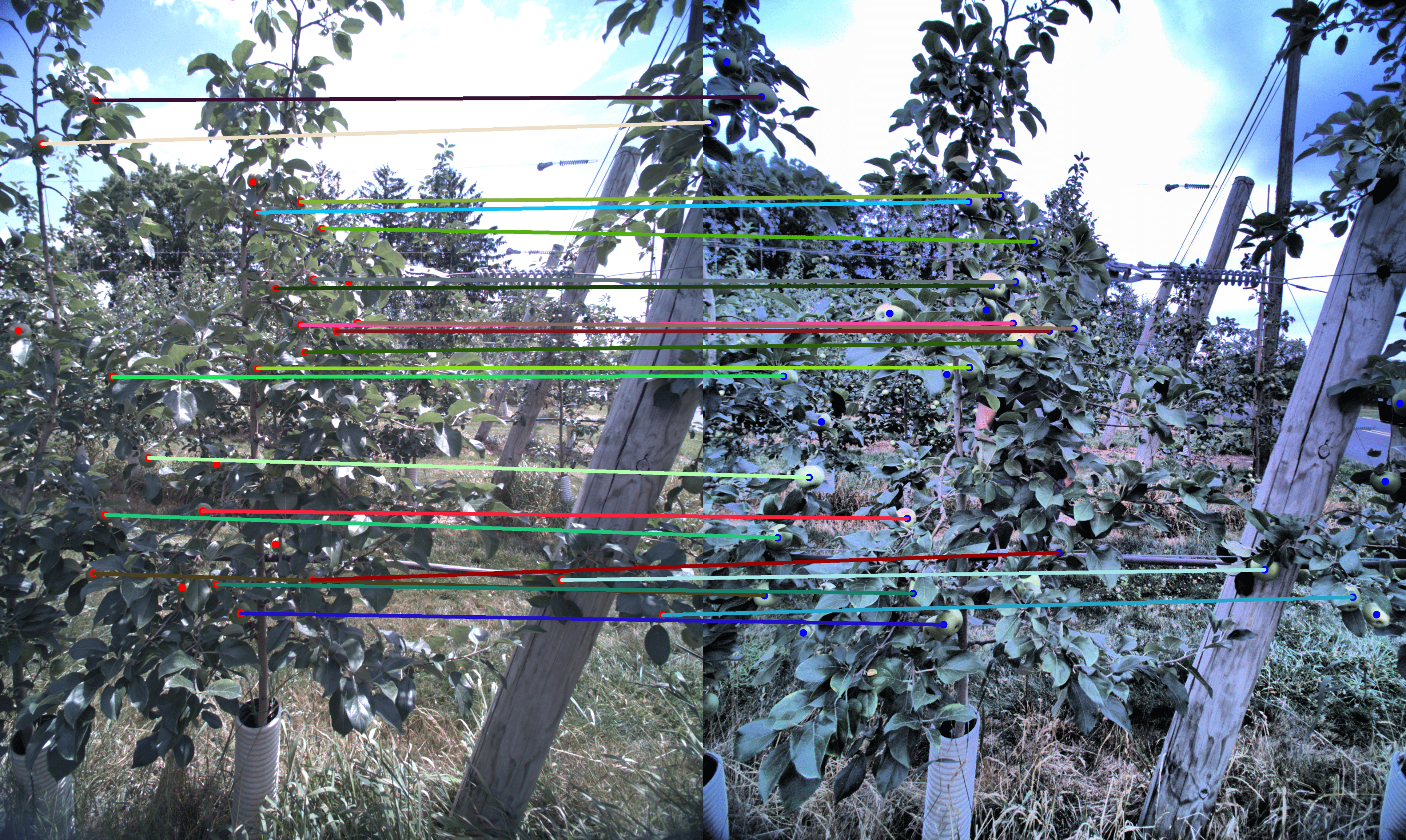}
    \end{subfigure}
    \begin{subfigure}[t]{0.31\textwidth}
        \centering          
            \includegraphics[width=1.0\textwidth]{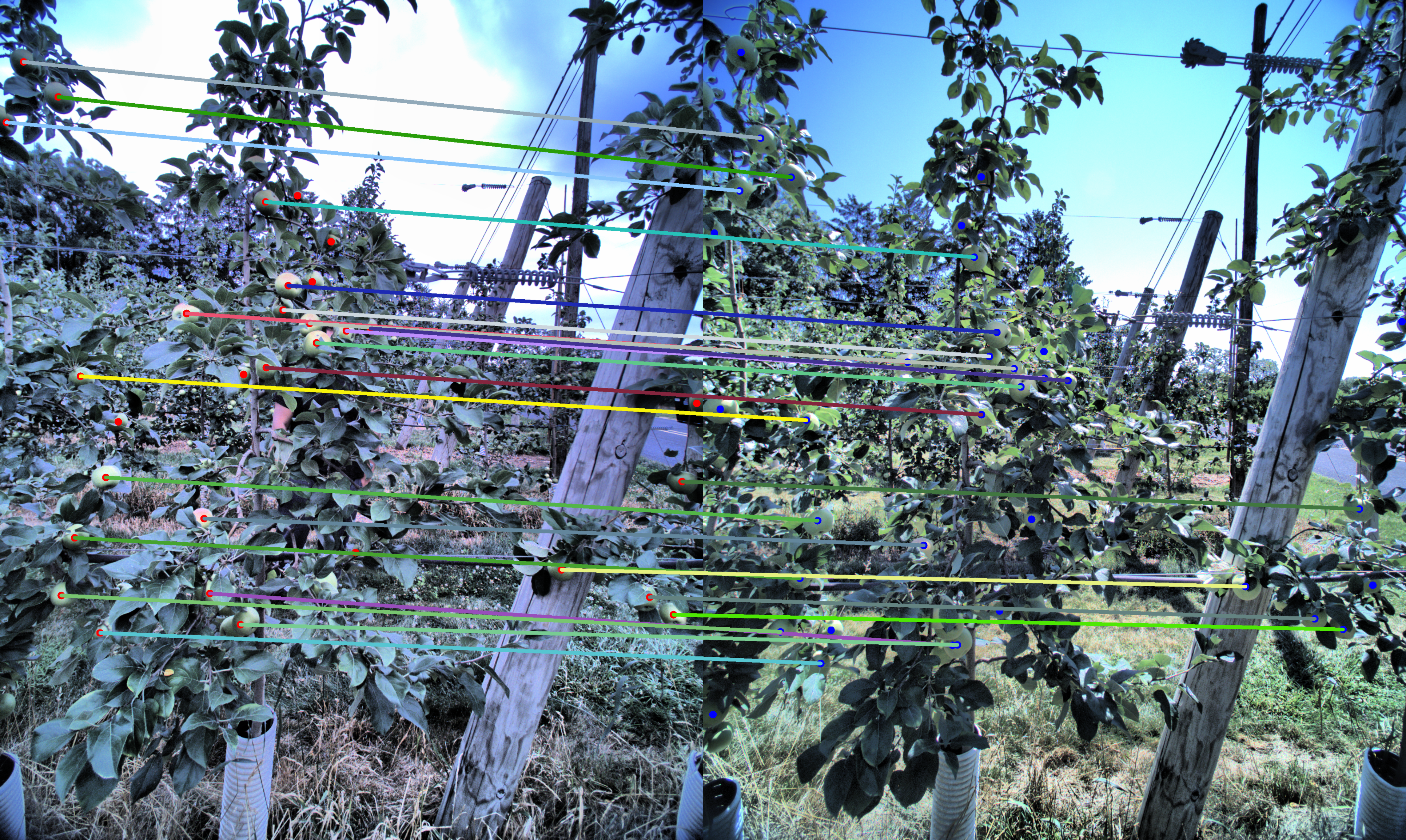}
    \end{subfigure}
    \begin{subfigure}[t]{0.31\textwidth}
        \centering          
            \includegraphics[width=1.0\textwidth]{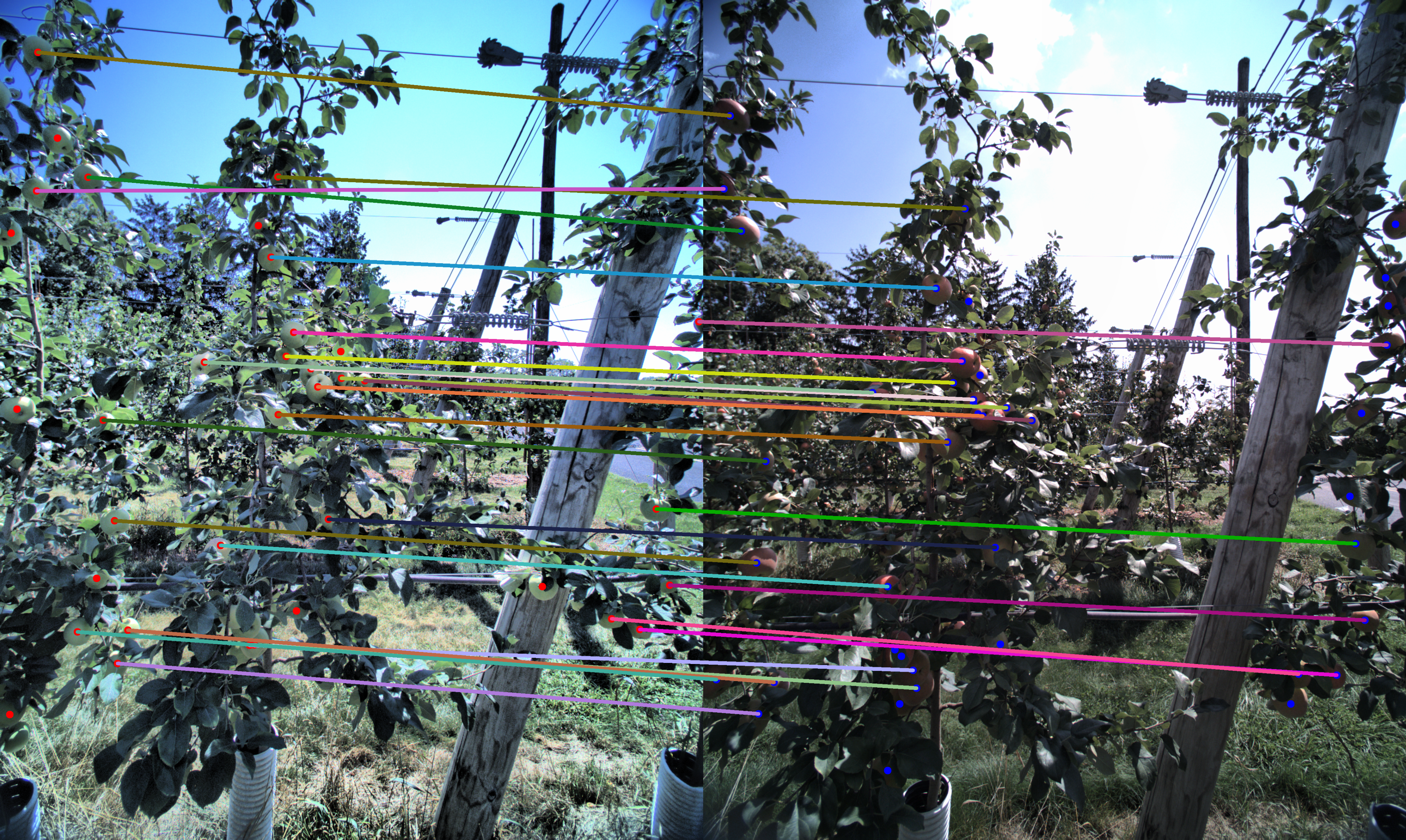}
    \end{subfigure}
    \caption{\textit{4D Fruit Matching}. \textit{\uline{The left panel}} illustrates the fruit matching between June 12th and July 4th, \textit{\uline{the middle panel}} between July 4th and July 19th, and \textit{\uline{the right panel}} between July 19th and August 13th. Red dots are fruits detected in the image frame from one time session. Blue dots are the fruits detected in the image frame from another time session. Matched fruits are linked with a line. 
    }
    \vspace{-0.2in}
\label{fig:fruit-cross-season-tracking-and-matching-visualization}
\end{figure*} 

To get the 4D fruit association ground truth, we manually match the fruits from five trees using images for each of the three consecutive time sessions—June 12th to July 4th, July 4th to July 19th, and July 19th to August 13th. We exclude a minority (less than 5 percent) of fruits that are ambiguous to match for humans. The 4D data association is evaluated based on two metrics: precision and F1 score. Higher precision means fewer incorrect fruit matches, while higher recall indicates that fewer correct matches were missed. 
However, F1 score is reported instead of recall for more meaningful evaluation because of two reasons: 1) Recall in our experiment is also affected by the matching range (eg. fruits from different time sessions that are 0.3 m away are not considered for matching.) A larger matching range tends to give a higher recall. 2) Failure to match fruits that should be matched usually has minimal effect on fruit growth monitoring. We usually have enough samples/matches to estimate the fruit growth rate, while falsely matching two fruits would lead to errors.
Precision and F1 scores are calculated from the number of true positives (TP), false positives (FP), and false negatives (FN). We define TP to be correctly matched fruits, FP to be incorrectly matched fruits, FN to be the fruits that should be matched but are predicted to have no available fruits to match with.

Refer to \cref{fig:precision} for the results on our approaches against the two baselines. The first baseline is to match fruits based on 3D positions only and associate them using the Hungarian assignment algorithm. The second baseline is a modified version of the algorithm proposed by Riccardi et al. in \cite{fruit_tracking_hign_precisoion_cloud}, where a histogram descriptor is designed to perform Hungarian assignment. Since the code of \cite{fruit_tracking_hign_precisoion_cloud} is not open-sourced, we re-implemented the algorithm and made the following modifications: 1) normalized the position cost and the cost based on histogram descriptors for them to be on the same scale which gives better performance than without normalization; 2) we excluded the use of fruit radius in the cost function as radius estimation of occluded fruits is not reliable in our case. Similarly, we used Optuna\cite{optuna_2019} on a subset of the fruits to tune the hyperparameters required by the baseline method. The average precision of our approach over the 4 time sessions is 93.5\%. The average F1 score is 92.1\%. Our approach improves the 4D fruit association by 23.7\% in precision and 11.9\% in F1 score in comparison with the second baseline.  

We conduct an ablation study to assess the contribution of different costs to 4D fruit association. Refer to \cref{fig:precision} for quantitative results. Combining position and visual information achieves comparable results to baselines. The inclusion of topology information further enhances matching precision by reducing false positives. However, the topology cost could mistakenly reject fruits that should be matched. We analyze the handling of FP by the topology cost to demonstrate the effectiveness: 43.2\% of the reduced FP are correctly rejected, 28.4\% are corrected to be TP, but 28.4\% of the reduced FP are incorrectly rejected when they should be TP. 
Furthermore, we compare the F1 scores of our methods against two baseline approaches by simulating real-world challenges such as occlusion, missed detections, and the emergence of new fruits, achieved by removing a percentage of fruits from the input to the 4D association method. As shown in \cref{fig:precision_with_removal}, the proposed 4D fruit association method consistently outperforms baseline methods, even when up to 50\% of the fruits are removed, highlighting its superior reliability to occlusion and missing data.

\subsection{Fruit Change Tracking and Size Estimation}
\label{subsec:results-fruit-size}

In \cref{fig:size_estimation}, we show that we can accurately estimate the size (diameter) of fruits up to a mean absolute error of 1.1 cm. 
For this experiment, we created a physical, real-world simulation of an apple tree in an orchard by manually attaching 19 apples to a similar tree. 
This was done since hand measuring and obtaining ground truth sizes of fruits in orchards is challenging due to difficulties in establishing a one-to-one association between ground truth and estimates. 
The fruits were hand-measured using a Vernier caliper to obtain their size ground truths. 
As seen from \cref{fig:size_estimation}, the size estimates closely follow the ground truth measurements. The errors induced in the size measurements are mainly due to the presence of background points such as leaves and stems in the segmentation mask of the fruits, which in turn inflates the extracted 3D fruit point clouds, leading to slightly larger size estimates compared to the ground truth. 
Further, we also show results on using our method to track the changes in the growth of fruits in uncontrolled orchards over multiple time sessions, as shown in the right panel of \cref{subfig:fruit-size-tracking-subfigure fruit size tracking}.

A visualization of our 4D tracking results on one example tree is shown in \cref{fig:fruit-cross-season-tracking-and-matching-visualization}. 
Spatio-temporal tracking of fruits allows for close monitoring of fruit size and appearance changes. 
An example of this can be seen in the left panel of \cref{fig:growth_vis} where each individual fruit is tracked and visualized across its growth progress. 
The resulting 4D metric-semantic map of the orchard, shown in \cref{fig:title-fig-4d-mapping}, illustrates how fruit sizes, canopy structure, and the overall appearance of both fruits and trees change over time.

%% file: tex/conclusion.tex
\section{Conclusion and Future Work}

This paper presents methods and datasets for 4D metric-semantic mapping of orchards. Real-world experiments demonstrate the effectiveness of the proposed approach, achieving a 3.1\% counting error in 3D fruit tracking and localization, a mean size estimation error of 1.10 cm for metric mapping, and a 23.7\% improvement in tracking precision over baseline methods. 
However, the proposed method also has some limitations. LiDAR sparsity limits the localization and size estimation of small fruits. In 4D fruit association, reliable reference point selection is enforced by a strict threshold, but outliers can still be chosen when computing topology cost. 

Furthermore, we aim to enable autonomous robots like UAVs for agricultural robotics. While a hand-carried sensor stack is currently used to map orchards, it can readily be mounted on robots, allowing them for autonomous data collection and orchard monitoring.